\documentclass[runningheads]{llncs}
\usepackage[T1]{fontenc}
\usepackage{array} % For column formatting
\usepackage{tabularray}
\usepackage{graphicx}  % For resizing tables and images
\usepackage{booktabs}  % For better horizontal lines in tables
\usepackage{caption}   % For customizing caption font size and style
\usepackage{amsmath}
\usepackage{amssymb}
\usepackage{wasysym}
\usepackage{caption}
\usepackage{pifont}
\usepackage{url}
\usepackage{wrapfig}
\usepackage{comment}
\usepackage{orcidlink}

\usepackage{color}
\begin{document}
%
%%%%%%%%%%%%%% Title %%%%%%%%%%%%%%%
\title{MuCo-KGC: Multi-Context-Aware Knowledge Graph Completion}

%%%%%%%%%%%%%%%%%%%%%%%%%%%%%%%%%%%

%\titlerunning{Abbreviated paper title}
% If the paper title is too long for the running head, you can set
% an abbreviated paper title here
%
\author{Haji Gul\inst{1a}\orcidlink{0000-0002-2227-6564} \and
 Abdul Ghani Naim\inst{1b}\orcidlink{0000-0002-7778-4961} \and Ajaz Ahmad Bhat\inst{1c}\orcidlink{0000-0002-6992-8224}
}

\authorrunning{H. Gul et al.}
% First names are abbreviated in the running head.
% If there are more than two authors, 'et al.' is used.

\institute{$^1$ School of Digital Science, Universiti Brunei Darussalam, Jalan Tungku Link, Gadong BE1410, Brunei Darussalam \\
\email{($^a$23h1710, $^b$ghani.naim, $^c$ajaz.bhat)@ubd.edu.bn}\\
%\url{https://sds.ubd.edu.bn/} 
%\and
%ABC Institute, Rupert-Karls-University Heidelberg, Heidelberg, Germany\\
%\email{\{abc,lncs\}@uni-heidelberg.de}
}
\maketitle              % typeset the header of the contribution
%%%%%%%%%%%%%%%%%%%%%%%% Abstract %%%%%%%%%%%%%%%%%%%%%%%%%%%%%%%%%%%%%%%%%%%
\begin{abstract}
%150-250
Knowledge Graph Completion (KGC) seeks to predict missing entities (e.g., heads or tails) or relationships in knowledge graphs (KGs), which often contain incomplete data. Traditional embedding-based methods, such as TransE and ComplEx, have improved tail entity prediction but struggle to generalize to unseen entities during testing. Textual-based models mitigate this issue by leveraging additional semantic context; however, their reliance on negative triplet sampling introduces high computational overhead, semantic inconsistencies, and data imbalance. Recent approaches, like KG-BERT, show promise but depend heavily on entity descriptions, which are often unavailable in KGs. Critically, existing methods overlook valuable structural information in the KG related to the entities and relationships. To address these challenges, we propose Multi-Context-Aware Knowledge Graph Completion (MuCo-KGC), a novel model that utilizes contextual information from linked entities and relations within the graph to predict tail entities. MuCo-KGC eliminates the need for entity descriptions and negative triplet sampling, significantly reducing computational complexity while enhancing performance. Our experiments on standard datasets, including FB15k-237, WN18RR, CoDEx-S, and CoDEx-M, demonstrate that MuCo-KGC outperforms state-of-the-art methods on three datasets. Notably, MuCo-KGC improves MRR on WN18RR, and CoDEx-S and CoDEx-M datasets by $1.63\%$, and $3.77\%$ and  $20.15\%$ respectively, demonstrating its effectiveness for KGC tasks.

\noindent\footnotetext{Accepted in PAKDD DSFA Special Session}
\end{abstract}
%%%%%%%%%%%% Introduction %%%%%%%%%%%%%%%%%%%%%%%%%%%%%%%%
\section{Introduction}
Knowledge Graphs (KGs) are structured representations of information about entities ${E}$ and relationships ${R}$, expressed as triplets ${T}$, each triplet consisting of a head, a relationship, and a tail $(h,r,t)$ \cite{bollacker2008freebase}. Integrated with deep learning and big data, KGs are increasingly being utilized in intelligent systems for tasks like question-answering, recommendation, and natural language processing \cite{zhu2024llms,miller1995wordnet}. KGs, despite their effectiveness, are often incomplete, missing data in the form of entities (heads and tails) or relationships between the entities, which limits their potential, as gaps in knowledge reduce the performance of downstream applications \cite{zhu2024llms}.  To address this, the Knowledge Graph Completion (KGC) or Link Prediction problem aims at predicting these missing entities ${E}$ and their interrelationships, ${R}$  thereby enhancing KG density \cite{yao2019kgbert}. In this work, we specifically focus on tail entity prediction, which involves inferring the missing tails in incomplete triplets\cite{yang2015distmult}. Tail prediction is generally more challenging than relation prediction due to the higher diversity and volume of unique entities compared to relationships set in most KGs. 
\par
KGC methods fall into three main categories: embedding-based, text-based, and LLM-based approaches. Embedding-based methods, such as DistMult \cite{yang2015distmult}, represent entities and relations as low-dimensional vectors but fail with unseen entities or relations during testing \cite{zhu2024llms}. These models also use fixed representations, ignoring context-specific variations \cite{yao2019kgbert}. Text-based models, like StAR \cite{wang2021StAR} and KG-BERT \cite{yao2019kgbert}, incorporate neighborhood and semantic information, improving predictions. However, they face scalability challenges, rely on expensive techniques like random walks, and often require entity descriptions, which are unavailable in many KGs. LLM-based approaches improve predictions through tailored prompting \cite{zhang2020pretrainkgc}, while recent works like DIFT \cite{liu2024finetuning} fine-tune generative LLMs, e.g., LLaMA-2 after combining these with embedding models. Despite these advances, LLM-based methods struggle to efficiently integrate large-scale KG facts, are domain-limited, and incur high computational costs.
\par
These challenges underscore the necessity for a scalable, description independent KGC framework that fully exploits KG structural properties and contextual information to enhance performance across diverse datasets. To address these gaps, we present \textbf{Multi-Context-Aware KGC (MuCo-KGC)}, a novel approach that extracts and integrates contextual information about \textit{entities} and \textit{relationships} associated with the head entity and query relation. This context is incorporated into a BERT-based architecture to improve tail entity prediction. Comprehensive experiments on multiple benchmark datasets demonstrate that MuCo-KGC achieves state-of-the-art performance in Mean Reciprocal Rank (MRR) and Hit@K metrics (specifically for $k = 1, 3$). To summarize, this study delivers the following \textbf{key contributions} to KGC domain:
\begin{itemize}
    \item \textbf{Contextualized Tail Prediction}: We propose MuCo-KGC, which leverages head entity and relationship contexts---structural KG features overlooked by prior models---to significantly improve predictive performance.
    \item \textbf{Description-Independence}: Unlike prior approaches, MuCo-KGC does not require entity descriptions, making it applicable to KG's with sparse or missing metadata.
    \item \textbf{Negative Sample-Free Training}: By utilizing cross-entropy loss, MuCo-KGC eliminates the need for negative sampling, improving training efficiency and robustness.
    \item \textbf{Empirical Validation}: Extensive experiments on various benchmark datasets demonstrate that MuCo-KGC reliably excels in tail entity prediction.
\end{itemize}

%%%%%%%%%%%%%%%%%%%%%%%%%%%%%% Related Work %%%%%%%%%%%%%%%%%%%%%%%%%%%%%%%%%%%%%%%%%%%%%%%
\section{Related Work}
The existing literature provides multiple KGC methods that can be broadly classified into two that have gained considerable attention. Knowledge Graph Embedding (KGE) and Language Model-Based Approaches.
\par   
KGE capture a graph's underlying semantics by representing entities and relationships as continuous vectors. These can further categorize into translation distance-based methods and semantic or textual matching-based methods. The first group includes TransE\cite{bordes2013trans}, RotatE \cite{sun2019rotate}, and others approaches \cite{Nickel2011rescal}, which train a scoring function to measure the distance between entities, modeling interactions as translation operations. In contrast, semantic or textual matching methods, such as DistMult \cite{yang2015distmult}, rely on scoring functions that compute the similarity between entities and relationships. These methods identify the most likely tail entity during testing by evaluating scores across all entities that share a specific head and relationship.
\par
Translation-based methods leverage relationships to facilitate transformations between entities $(h, t)$, enabling triplet scoring. TransE, a foundational model in this category, embeds entities and relationships in a shared $d$-dimensional vector space and computes triplet scores using $f_{\text{TransE}}(h, r, t) = -||h + r - t||$, where $h$, $r$, and $t \in \mathbb{R}^d$ \cite{bordes2013trans}. It assumes $h + r \cong t$, effectively modeling simple relational patterns but failing with complex relationships such as $1-N$, $N-1$, and $N-N$ mappings. RotatE addresses this limitation by embedding entities and relationships in a complex vector space $\mathbb{C}^d$ and employing rotational transformations, defined as $f_{\text{RotatE}}(h, r, t) = -||h \circ r - t||$ \cite{sun2019rotate}. Extending this, HAKE adopts a polar coordinate system to capture semantic hierarchies, improving its ability to model hierarchical relations \cite{zhang2020hake}. In contrast, semantic-based methods like DistMult \cite{yang2015distmult} utilize matrix-based operations to measure entity-relation similarity. The scoring function $f_{\text{DistMult}}(h, r, t) = h^\top \text{diag}(r) t$, with $h$, $r$, and $t \in \mathbb{R}^d$, excels at modeling compositional semantics. However, its symmetric nature limits its ability to handle asymmetric relationships, reducing its accuracy in predicting missing knowledge graph information \cite{zhang2020hake}.

\par
Recent studies have explored text-based models for KGC, with {KG-BERT} \cite{yao2019kgbert} being a pioneering approach. KG-BERT models knowledge graph triplets as ordered sequences and employs a binary cross-entropy loss to classify their validity. While effective in triplet classification, it under-performs in some ranking metrics, especially on Hits@1 and Hits@3, compared to state-of-the-art methods. Building on this, {MEM-KGC} \cite{choi2021memkgc} adopts a masked language model approach, masking the tail entity in a triplet and predicting it using relational and textual descriptions. Although MEM-KGC achieves notable performance gains, its dependence on textual descriptions limits its applicability to KGs with sparse or absent entity descriptions. These text-based models, despite their advancements, struggle to generalize effectively across diverse knowledge graph datasets that lack comprehensive descriptive metadata.
\par
Recent hybrid methods seek to combine the strengths of large language models (LLMs), prompt engineering  \cite{wei2024kicgpt} and traditional KGE techniques. For example, Pretrain-KGC \cite{zhang2020pretrainkgc} integrates LLMs with classical models like TransE and RotatE by initializing embeddings in a shared vector space, subsequently refining them through iterative optimization to improve loss detection. Similarly, DIFT \cite{liu2024finetuning} employs fine-tuned LLaMA-2-7B models augmented with a knowledge adaptation module, enabling enhanced reasoning over KG-specific domains. Despite achieving state-of-the-art results, DIFT’s dependence on computationally intensive LLMs like LLaMA-2-7B limits its practicality in resource-constrained settings.
\par
To summarize, existing KGC methodologies exhibit notable limitations. Embedding based models, such as TransE and RotatE, fail to generalize to unseen entities and relations due to their dependence on pre-defined embeddings, limiting adaptability to new knowledge graph structures. Text-based and LLM-based approaches, including KG-BERT and MEM-KGC, heavily depend on descriptive entity information, which is often sparse or inconsistent across datasets. Many embedding methods as well as some  text-based methods like  SimKGC \cite{wang2022simkgc} rely on extensive negative sampling during training, leading to high computational costs, especially on large datasets. Furthermore, LLM-based methods like KICGPT and DIFT introduce additional complexity through prompt engineering and fine-tuning, further increasing their computational and resource demands.
\par
A critical limitation of existing KGC methods is their insufficient exploitation of the structural properties of KG's. Many approaches overlook the rich contextual information embedded in the graph's structure, such as neighboring entities and relationships tied to a given head entity and query relation. Key structural components that can provide valuable context include \textit{head context-entities} (neighboring entities), \textit{head context-relations} (neighboring relations), and the \textit{relationship context} (information specific to the query relation).
\par
The proposed MuCo-KGC model addresses this gap by integrating these three structural dimensions to construct a comprehensive context. By leveraging neighboring entity information, surrounding relational details, and relationship-specific context, MuCo-KGC captures a deeper understanding of the graph’s topology. This integration of textual and structural information represents a significant advancement over existing methods, enabling improved prediction accuracy and enhanced generalization. The following section details the design and implementation of the MuCo-KGC model.

%%%%%%%%%%%%%%%%%%%%%%%%%%%%% Methodology %%%%%%%%%%%%%%%%%%%%%%%%%%%%%%%%%%%%%%%%%
\section{Methodology}
This section describes the MuCo-KGC technique, which predicts tail entities given a head entity and a relationship from a knowledge graph. Before illustrating the model, a brief description of the problem is provided. 
\par
\textbf{Problem Statement}: Consider a knowledge graph $G(E, R, T)$ as a collection of triplets $T$, where each triplet is given as ${(h, r, t)}$. Here, $ h ~\in ~ E ~ $ is the head entity, $ t ~ \in ~ E$ is the tail entity, and $r ~ \in ~ R$ represents the relationship between them. Given an incomplete triple $(h ~ , ~ r ~, ~ ?)$, MuCo-KGC model predicts the missing tail $t$ (represented by ? in the incomplete triple). 
\begin{figure}
  \includegraphics[width= 1\linewidth]{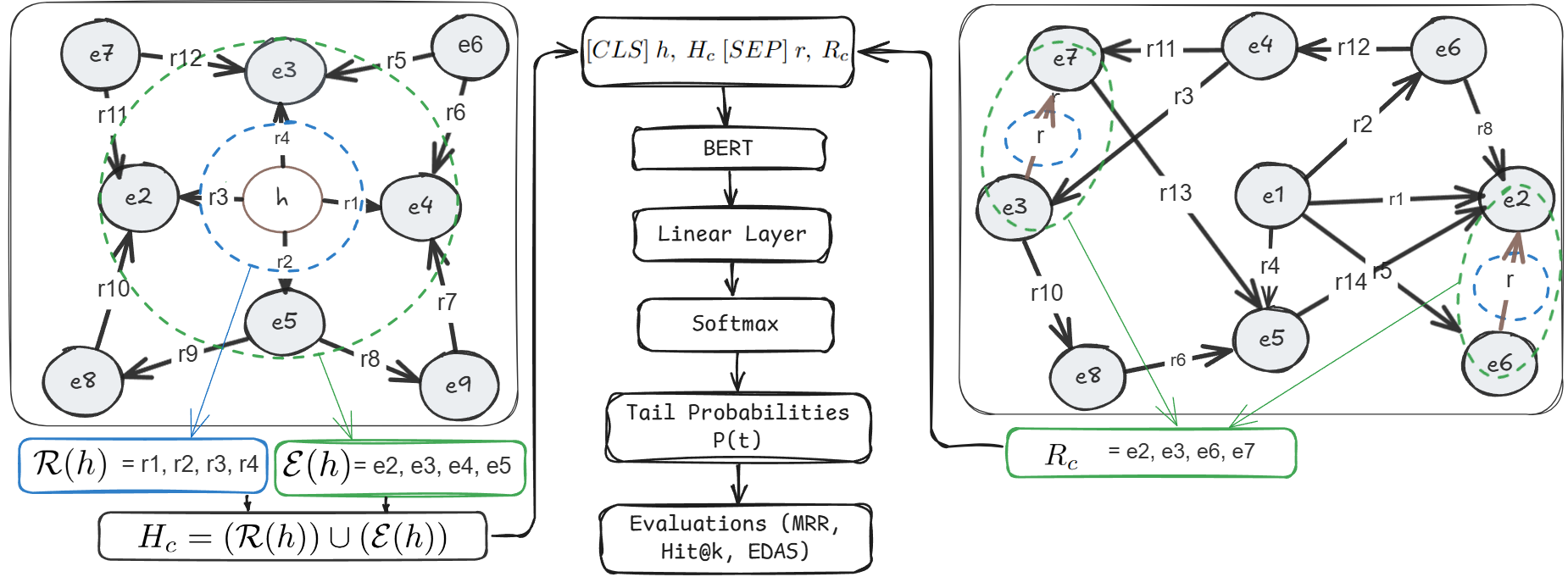} 
  \captionsetup{font=small} 
  \caption {A concise overview of the MuCo-KGC model pipeline for predicting the tail entity, given a head entity $h$ and a relationship $r$. The box on the left illustrates the calculation of head context $H_c$. $H_c$ is formed as a union of $\mathcal{R}(h)$ and $\mathcal{E}(h)$. Here, $\mathcal{R}(h)$ is the set of all relations ($r1$, $r2$, $r3$ and $r4$) involving the head entity $h$, while $\mathcal{E}(h)$ is the set of all neighboring entities ($e2$, $e3$, $e4$, and $e5$) directly related to $h$.  The box on the right shows the calculation of relationship context $R_c$. $R_c$ comprises the set of all entities ($e3$, $e7$, $e2$, and $e6$) associated via relationship $r$. These contextual features --- $H_c$ and $R_c$ --- alongside $h$ and $r$ are then fed as input to the BERT model as depicted in the middle of the figure.  The BERT model, combined with a linear classifier and softmax, generates probabilities for tail entities.}
  \label{f:kgc-ca-quick}
\end{figure}

To achieve accurate tail prediction, the proposed MuCo-KGC method extracts contextual information from the knowledge graph surrounding the head entity and the relationship in question, and leverages this context within a BERT model to learn and generate precise predictions. Figure \ref{f:kgc-ca-quick} provides a detail overview of the MuCo-KGC model. Given a head  $h$ and a relationship $r$, our model predicts the tail entity $t$ , in the following steps: 
\par

%\textbf{Step 1. Extract Head Context} ${H}_c$:
\textbf{Extract Head Context} ${H}_c$:
To extract the contextual information for the head, i.e. ${H}_c$, we first identify the relationships $r$ that are associated with the head entity $h$, i.e., the relations neighborhood $\mathcal{R}(h)$.  If  $k$ relationships are associated with the head $h$ from  the set $R$ of all relationships $r_i$ in the graph $G$, then:
\begin{equation}
 \mathcal{R}(h) = A_{i=1}^{k}\left(\{r_i \mid (h, r_i, e_j) \in T, \, e_j \in {E}\}\right)
\end{equation}
\par

where $ A(\cdot) $ is some aggregator function (in our case, it is the concatenation operation $ \| $ ),  $ T $ is the set of training triplets, $ E $ is the set of all entities and $ r_i $ represents each relation associated with $h$.
\par

Next, we find the entities $e$ that are neighbors (i.e., have a direct connection) with the head entity $h$, i.e., entity neighborhood $ \mathcal{E}(h)$ using the identified relationships in $ \mathcal{R}(h)$. Assuming  $m$ neighbor entities, $ \mathcal{E}(h)$ is expressed as:
\begin{equation}
     \mathcal{E}(h) = A_{i=1}^{m}\left(\{e_i \mid (h, r_j, e_i) \in T, \, r_j \in {R}\}\right)
\end{equation}
\par

where $ R $ is the set of all relations and $ e_i $ represents each entity directly associated with $ h $ The head context $H_c$ is then calculated as the union of the connected relations $ \mathcal{R}(h)$ and the neighbor entities $ \mathcal{E}(h)$, as shown below in Equation \ref{eq:hc}.\\
\begin{equation}
{H}_c = \left( \mathcal{R} (h) \right) \cup \left( \mathcal{E} (h) \right)
\label{eq:hc}
\end{equation}

Conceptually, the head context ${H}_c$  represents the entity-centric local knowledge. It pulls out:
\begin{itemize}
    \item Entities directly linked to the head entity $ \mathcal{E}(h)$: This captures local neighborhood information, essentially providing a "snapshot" of the immediate surroundings of the head entity in the graph. This structural information helps the model identify how the head entity $h$ is positioned within its local neighborhood (e.g., is it a hub, part of a chain, etc.).
    \item Relations associated with these neighboring entities $ \mathcal{R}(h)$: By including the relationships between the head entity and its neighbors, we bring in a semantic understanding of the types of relationships  the entity $h$ is generally involved in.
\end{itemize}
\par

Therefore, the head context's role is to provide \textbf{specificity}. By narrowing the focus to the immediate surroundings of the head entity $h$, it helps the model differentiate between plausible tail entities based on the specific neighborhood structure. 
\par

%\textbf{Step 2. Extract Relationship Context} ${R}_c$:
\textbf{Extract Relationship Context} ${R}_c$: To acquire the relationship context ${R}_c$, we identify all the entities associated with the operational relationship $r$ in the knowledge Graph $G$. $R_c$ is given below, where $ e_i, e_j $ are entities in $ E $ connected by the relation $ r $.
\begin{equation}
    R_c = A_{i,j=1}^{l}\left(\{e_i, e_j \mid (e_i, r, e_j) \in T\}\right)
    \label{eq:rc}
\end{equation}
\par

Conceptually, the relation context ${R}_c$ represents relation-centric global knowledge. It pulls out all entities connected via the given relation $r$.  Instead of focusing on the head entity’s neighborhood, this provides a global perspective on the nature of the relation $r$ itself and the patterns or clusters involving the relation $r$. This helps the model learn global patterns about how this relation typically connects entities across the entire KG (e.g., relation \textit{capital\_of} connects countries and cities). Relation context $R_c$ also provides the model with the knowledge of the distribution or variability of the manifestation of the relation $r$ in the KG. This helps the model refine predictions by learning how the relation $r$ operates beyond the local context of the head entity. In other words, the relation context's role is to provide \textbf{generalization}.  By considering global patterns of the relation, it acts as a regularizer, ensuring that the model aligns with broader relational constraints in the KG. Together, the head and relation contexts allow the model to leverage local specificity (head context) and ensure global consistency (relation context). This innovation solves the "description sparsity" problem in dealing with datasets without descriptions. 
\par

%\textbf{ Step 3. Prepare Input Sequence for BERT Classifier}:
\textbf{Prepare Input Sequence for BERT Classifier}: The contextual information extracted in the above steps forms the input to BERT. Specifically, the input sequence contains h, ${H}_c$ from  Equation \ref{eq:hc}, $r$, and ${R}_c$ from Equation \ref{eq:rc}, as shown below where [CLS] is BERT's classifier token and [SEP] is the separator token.
\begin{equation}
    \text{Input Sequence}   =[CLS] ~ h, ~ {H}_c ~ [SEP] ~ r, ~{R}_c
\end{equation}
\par

%\textbf{Step 4. Predict and train with BERT Classifier}:
\textbf{Predict and train with BERT Classifier}:
A classification layer is added on top of the BERT model, which aims to classify all the entities from Entity set ${E}$. Once the BERT classifier receives the input, it processes it through various transformer layers, provides a contextualized representation of each token, and uses that to classify the input.  The classifier model predicts the tail entity by employing a softmax function over the output pooled representation to calculate the probability for all the available tail entities. The input-output description of the model is given as: 
\begin{equation}
    \text{(BERT Output) } =  \text{ BERT(Input Sequence)}
\end{equation}
\begin{equation}
    P(t \mid h, r) = \text{softmax}(W \cdot \text{ BERT(Input Sequence)})
\end{equation}
\par

where $ W $ is a learnable weight matrix. Putting the above equations together, the MuCo-KGC model can be expressed as:
\begin{equation}
    \text{MuCo-KGC}(t \mid h, r) = \text{softmax}(W \cdot \text{ BERT}(h,{H}_c, r,{R}_c))
\end{equation}
\par

The MuCo-KGC model is trained using cross-entropy loss, which compares the probability distribution of the predicted label with the true label for the tail entity. The cross-entropy loss is given by:
\setlength{\belowdisplayskip}{0pt}
\begin{equation}
    {L} = - \sum_{i=1}^{N} y_i \log P(t_i \mid h, r)g
\end{equation}
\par

In this equation, the one-hot encoded true label for the tail object $ t_i $ is indicated as $ y_i $. The predicted probability for the true tail entity could be denoted as $ P(t_i \mid h, r) $, where $h$ is the head and $r$ is the relation. \\

\noindent
\textbf{Computational Complexity}: The complexity of the three operations over the entire graph is as follows:

\textit{Relational Neighborhood Aggregation} ($\mathcal{R}(h) $):
For node $ h $, the complexity is $ O(k) $, where $ k $ is the number of relations connected to $ h $. Over all nodes, $ O\left(\sum_{h \in E} k_h\right) = O(|T|) $, where $ |T| $ is the total number of triplets in the graph. 

\textit{Neighbor Entity Aggregation} ($ \mathcal{E}(h)$):
For a single node $ h $, the complexity is $ O(m) $, where $ m $ is the number of neighboring entities. Aggregating over all nodes, $
O\left(\sum_{h \in E} m_h\right) = O(|T|).$

\textit{Relation Context Aggregation} ($R_c(r) $):
For a single relation $ r $, the complexity is $ O(l) $, where $ l $ is the number of triplets involving $ r $. Across all relations, 
$O\left(\sum_{r \in R} l_r\right) = O(|T|).$

Therefore, the \textbf{overall complexity} for context calculation is $O(3 \cdot |T|) $ staying linear in $ |T| $ ensuring computational efficiency, even for large-scale graphs. Note that context is calculated once only at training. Consequently, our model time complexity at training is  $ O(3 \cdot |T|) + O(|T| \cdot |BERT|) $ and that at inference is only $O(|T| \cdot |BERT|) $, where $ |BERT| $ is the complexity of a BERT forward pass. This is efficient compared to similar state-of-the-art text-based BERT methods like SimKGC\cite{wang2022simkgc} with training and inference complexities of $ O(2 \cdot N \cdot |T| \cdot |BERT|) $ and $ O(2 \cdot |T| \cdot |BERT|) $ respectively at the least, where $ N \geq 3 $ is the number of negative samples for each triple.

\section{Experiments}
\textbf{Datasets:} We evaluate the proposed MuCo-KGC model on four widely used knowledge graph (KG) datasets:

\begin{itemize}
    \item \textbf{FB15k-237} \cite{bollacker2008freebase}: An updated version of FB15k with inverse triplets removed to increase difficulty. It consists of 14,541 entities, 237 relations, 272,115 training, 17,535 validation, and 20,466 test triplets.
    \item \textbf{WN18RR} \cite{miller1995wordnet}: A subset of WN18, where reverse triplets are removed for increased complexity. The dataset includes 40,943 entities, 11 relations, 86,835 training, 2,924 validation, and 2,824 test triplets.
    \item \textbf{CoDEx-S} \cite{safavi2020codex}: A multi-domain KG sourced from Wikidata, featuring 2,034 entities, 42 relations, 32,888 training, 3,654 validation, and 3,656 test triplets.
    \item \textbf{CoDEx-M} \cite{safavi2020codex}: A medium-sized version of CoDEx KG with 17,050 entities, 51 relations, and 185,584 triples, designed for complex KGC tasks such as link prediction and triple classification.
\end{itemize}

\noindent
\textbf{Hyperparameters:} The experiments used a batch size of 16, a learning rate of 5e-5, Adam optimizer, and cross-entropy loss. All experiments were conducted on an NVIDIA GeForce RTX 3090 GPU (24 GB memory). Training was stopped when evaluation metrics stabilized to three decimal places, typically within 30 epochs.
\par
\noindent
\textbf{Evaluation:} The proposed method and existing approaches are evaluated using standard KGC metrics, including MRR and Hits@k, defined in Equation \ref{eq:eva}:
\begin{equation}
\label{eq:eva}
\text{MRR} = \frac{1}{N} \sum_{i=1}^{N} \frac{1}{\text{rank}_i}, \quad  
\text{Hits@k} = \frac{1}{N} \sum_{i=1}^{N} \mathbf{1}(\text{rank}_i \leq k)
\end{equation}
where $\text{rank}_i$ denotes the rank of the correct entity in the sorted list of predicted scores for the $i$-th triplet, and $\mathbf{1}(\text{rank}_i \leq k)$ equals 1 if the correct entity is among the top $k$ predictions, and 0 otherwise.

%%%%%%%%%%%%%%%%%%%%%%%%%%%%%%%%%%%%%%%%%%%%%%%%%%
\subsection{Results}

%%%%%%%%%%%%% Table 1%%%%%%%%%%%%%%%%%%%%%%%%%%%%%%%%%
\begin{table*}[h!]
\centering
\captionsetup{font=small}
\caption{Tail prediction results on FB15k-237 and WN18RR datasets. The best result for each metric is in \textbf{boldface}, and the second-best is underlined. Results marked with the $\Circle$ symbol are from \cite{wei2024kicgpt}, and those marked with the $\Box$ symbol are from \cite{yao2019kgbert}.}
\label{tab:results}
\renewcommand{\arraystretch}{1.1} % Increase row height
\resizebox{\textwidth}{!}{%
\begin{tabular}{lp{1.5cm}p{1.5cm}p{1.5cm}p{1.5cm}p{1.5cm}p{1.5cm}p{1.5cm}p{1.5cm}}
\toprule
\textbf{Dataset} & \multicolumn{4}{c}{\textbf{FB15k-237}} & \multicolumn{4}{c}{\textbf{WN18RR}} \\ 
\cmidrule(lr){2-5} \cmidrule(lr){6-9}
\textbf{Methods} & \textbf{MRR $\uparrow$} & \textbf{Hits@1 $\uparrow$} & \textbf{Hits@3 $\uparrow$} & \textbf{Hits@10 $\uparrow$} & \textbf{MRR $\uparrow$} & \textbf{Hits@1 $\uparrow$} & \textbf{Hits@3 $\uparrow$} & \textbf{Hits@10 $\uparrow$} \\
\midrule
\textbf{\textit{Embedding-Based Methods}} & & & & & & & & \\
RESCAL \cite{Nickel2011rescal} $\Circle$      & 0.356 & 0.266 & 0.390 & 0.535 & 0.467 & 0.439 & 0.478 & 0.516 \\  
TransE \cite{wei2024kicgpt}   $\Circle$       & 0.279 & 0.198 & 0.376 & 0.441 & 0.243 & 0.043 & 0.441 & 0.532 \\  
TuckER \cite{16}  $\Circle$                   & 0.358 & 0.266 & 0.394 & 0.544 & 0.470 & 0.443 & 0.482 & 0.526 \\  
ComplEx \cite{11}  $\Circle$                  & 0.247 & 0.158 & 0.275 & 0.428 & 0.440 & 0.410 & 0.460 & 0.510 \\  
ConvE \cite{Dettmers2018ConvE}                & 0.310 & 0.239 & 0.350 & 0.491 & 0.460 & 0.390 & 0.430 & 0.480 \\  
DistMult \cite{yang2015distmult}  $\Circle$   & 0.241 & 0.155 & 0.263 & 0.419 & 0.430 & 0.390 & 0.440 & 0.490 \\  
RotatE \cite{sun2019rotate}  $\Circle$        & 0.338 & 0.241 & 0.375 & 0.533 & 0.476 & 0.428 & 0.492 & 0.571 \\  
CompGCN \cite{Vashishth2020CompGCN} $\Circle$ & 0.355 & 0.264 & 0.390 & 0.535 & 0.479 & 0.443 & 0.494 & 0.546 \\  
HittER \cite{wei2024kicgpt}  $\Circle$        & 0.344 & 0.246 & 0.380 & 0.535 & 0.496 & 0.449 & 0.514 & 0.586 \\  
HAKE \cite{zhang2020hake}  $\Circle$          & 0.346 & 0.250 & 0.381 & 0.542 & 0.497 & 0.452 & 0.516 & 0.582 \\  
BiQUE \cite{guo2021bique}                     & 0.365 & 0.270 & 0.401 & 0.555 & 0.504 & 0.459 & 0.519 & 0.588 \\  
\midrule
\textbf{\textit{Text-and Description-Based Methods}} & & & & & & & & \\
Pretrain-KGE \cite{zhang2020pretrainkgc} $\Circle$ & 0.332 &   -   &  -   & - & 0.235 & - & - & - \\  
StAR \cite{wang2021StAR}  $\Circle$           & 0.263 & 0.171 & 0.287 & 0.452 & 0.364 & 0.222 & 0.436 & 0.647 \\  
MEM-KGC (w/o EP) \cite{choi2021memkgc} $\Circle$ & 0.339 & 0.249 & 0.372 & 0.522 & 0.533 & 0.473 & 0.570 & 0.636 \\  
MEM-KGC (w/ EP) \cite{choi2021memkgc} $\Circle$  & 0.346 & 0.253 & 0.381 & 0.531 & 0.557 & 0.475 & 0.604 & 0.704 \\  
SimKGC \cite{wang2022simkgc} $\Circle$        & 0.333 & 0.246 & 0.363 & 0.510 & 0.671 & 0.585 & \textbf{0.731} & \textbf{0.817} \\  
\midrule
\textbf{\textit{LLM-Based Methods}}           & & & & & & & & \\
ChatGPTzero-shot \cite{zhu2024llms}  $\Box$   & - & 0.237  & - & - & - & 0.190 & - & - \\  
ChatGPTone-shot \cite{zhu2024llms}   $\Box$   & - & 0.267  & - & - & - & 0.212 & - & - \\  
DIFT \cite{liu2024finetuning}                 & \textbf{0.439} & \textbf{0.364} & \textbf{0.468} & \underline{0.586} & 0.617 & 0.569 & 0.638 & 0.708 \\  
KICGPT \cite{wei2024kicgpt}   $\Box$          & 0.412 & 0.321 & 0.448 & 0.554 & 0.549 & 0.474 & 0.585 & 0.641 \\  
\midrule
\textbf{\textit{GNN-Based Method}}            & & & & & & & & \\
NBFNet \cite{42}                              & \underline{0.415} & 0.321 & \underline{0.454} & \textbf{0.599} & 0.551 & 0.497 & 0.573 & 0.666 \\  
NNKGC \cite{22} $\Circle$                     & 0.338  & 0.252  & 0.365 & 0.515 & \underline{0.674} & \underline{0.596} & \underline{0.722} & \underline{0.812} \\  
\midrule
\textbf{\textit{Proposed}}                    & & & & & & & & \\
MuCo-KGC (H$_c$ Only)                         & 0.310 & 0.263 & 0.331 & 0.403 & 0.420 & 0.492 & 0.556 & 0.616 \\  
MuCo-KGC (R$_c$ Only)                         & 0.280  & 0.187 & 0.255 & 0.271 & 0.321 & 0.345 & 0.371 & 0.398 \\  
\textbf{MuCo-KGC}                             & 0.350 & \underline{0.322} & 0.399 & 0.462 & \textbf{0.685} & \textbf{0.637} & 0.687 & 0.737 \\  
\bottomrule
\end{tabular}%
}
\end{table*}

%%%%%%%%%%%%% Table 2 %%%%%%%%%%%%%%%%%%%%%%%%%%%%%%%%%
\begin{table*}[h!]
\centering
\captionsetup{font=small}
\caption{Tail prediction results on the CoDEx-S and CoDEx-M datasets. The best result for each metric is in \textbf{boldface}, and the second-best is underlined. Results are taken from \cite {zeb2024,song2023openfact}.}
\label{tab:results4}
\scriptsize % Decrease font size slightly less than \scriptsize
%\scriptsize
\renewcommand{\arraystretch}{1.1} % Increase row height
\resizebox{\textwidth}{!}{%
\begin{tabular}{lp{1.5cm}p{1.5cm}p{1.5cm}p{1.5cm}p{1.5cm}p{1.5cm}p{1.5cm}p{1.5cm}}
\toprule
\textbf{Dataset} & \multicolumn{4}{c}{\textbf{CoDEx-S}} & \multicolumn{4}{c}{\textbf{CoDEx-M}} \\ 
\cmidrule(lr){2-5} \cmidrule(lr){6-9}
\textbf{Methods} & \textbf{MRR $\uparrow$} & \textbf{Hits@1 $\uparrow$} & \textbf{Hits@3 $\uparrow$} & \textbf{Hits@10} & \textbf{MRR $\uparrow$} & \textbf{Hits@1 $\uparrow$} & \textbf{Hits@3 $\uparrow$} & \textbf{Hits@10} \\
\midrule
%\textbf{\textit{Embedding-Based Methods}} & & & & & & & & \\
RESCAL \cite{Nickel2011rescal}      & 0.40 & 0.293 & - & 0.623 & 0.317 & 0.244 & - & 0.456 \\  
TransE \cite{wei2024kicgpt} & 0.44 & 0.339 & - & 0.638 & 0.303 & 0.223 & - & 0.454 \\ 
TuckER \cite{16}      & 0.44 & 0.339 & - & 0.638 & .328 & 0.259 & - & 0.458 \\ 
OpenFact \cite{song2023openfact} & \underline{0.53} &\textbf{0.430}& - & \underline{0.736} & \underline{0.403} & \underline{0.313} & - & \underline{0.578} \\ % new
HoLE-RorL \cite{wang2022swift} &0.40 &0.292& - & 0.639 & 0.326& 0.246 & - & 0.475 \\ % new
ComplEx \cite{11}         & 0.46 & 0.370 & - & 0.646 & 0.337 & 0.262 & - & 0.476 \\ 
% for all below i will cite node co-occurance paper. 
ConvE \cite{Dettmers2018ConvE}           & 0.44 & 0.343 & - & 0.635 & 0.318 & 0.239 & - & 0.464 \\ 
RotL \cite{wang2021hyperbolic} & 0.408 & 0.292& - & 0.639 & 0.324 & {0.244}  &  - & 0.474 \\
CoPE \cite{zeb2024} & 0.446 & 0.350& - & 0.631 & 0.326 & 0.251  &  - & 0.466 \\
\textbf{MuCo-KGC} & \textbf{0.55} & \underline{0.423} & \textbf{0.647} & \textbf{0.764} & \textbf{0.478} & \textbf{0.392} & \textbf{0.523} & \textbf{0.600} \\  
\bottomrule
\end{tabular}%
}
\end{table*}

%%%%%%%%%% commneted out models for Codex-s and CoDEx-M datasets
%Dinov2 \cite{zhumultimodal} & 0.40 &0.358 &0.427 &0.473& 0.071& 0.022 & 0.0960 & 0.1398 \\ % new
%SimRGCN  & 0.42 & - & - & 0.647 & 0.322 & - & - & 0.475 \\ % \cite{nguyen2021quaternion} 
%SimQGNN    & 0.43 & - & - & 0.652 & 0.323 & - & - & 0.477 \\ %\cite{nguyen2021quaternion}
%R-GCN    & 0.27 & - & - & 0.533 & 0.124 & - & - & 0.241 \\  % \cite{dai2024granular}
%CompGCN  & 0.39 & - & - & 0.650 & 0.124 & -  &  - & 0.241 \\ % \cite{vashishth2020}
The experimental results, presented in Tables \ref{tab:results} and \ref{tab:results4}, provide a comparative analysis of MuCo-KGC against state-of-the-art models across FB15k-237, WN18RR, CoDEx-S, and CoDEx-M datasets. Performance is evaluated using standard metrics: Mean Reciprocal Rank (MRR), Hits@1, Hits@3, and Hits@10.

\noindent
\textbf{Performance on WN18RR and FB15k-237:} 
MuCo-KGC achieves state-of-the-art performance on WN18RR in two metrics, attaining an MRR of 0.685, surpassing the previous best, NNKGC (MRR = 0.674), and leading in Hits@1 (0.637). The results on Hits@3 (0.687), and Hits@10 (0.737) are third-best. These results underline its ability to capture intricate relational patterns. On FB15k-237, MuCo-KGC achieves competitive performance with an MRR of 0.350 and Hits@10 of 0.462, slightly behind methods like DIFT and NBFNet. This indicates potential for improvement in densely connected graphs.

\noindent
\textbf{Performance on CoDEx-S and CoDEx-M:} 
MuCo-KGC excels on both CoDEx datasets. On CoDEx-S, it achieves the highest MRR (0.550) and leads in Hits@3 (0.647) and Hits@10 (0.764). While Hits@1 (0.423) trails OpenFact (0.430), MuCo-KGC's overall ranking performance is superior. On CoDEx-M, it outperforms all baselines, with an MRR of 0.470 and the highest Hits@1 (0.392), Hits@3 (0.523), and Hits@10 (0.600), significantly surpassing OpenFact (MRR = 0.403) and ComplEx (MRR = 0.337).

\begin{wrapfigure}{r}{0.5\textwidth} % 'r' for right alignment, '0.4\textwidth' is the width of the table
\captionsetup{type=table} % Ensures the caption is labeled as Table
\caption{A comparison of model parameters.}
\label{tab:prc}
\centering
\begin{tabular}{|c|c|}
\hline
\textbf{Model} & \textbf{Number of parameters} \\ \hline
DIFT &  7 \textbf{B} \\ \hline
Sim-KGC & 220 \textbf{M} \\ \hline
MuCo-KGC & 110 \textbf{M} \\ \hline
\end{tabular}
\end{wrapfigure}

\noindent
\textbf{Insights and Hypothesis Validation:} 
MuCo-KGC outperforms traditional KG embedding models, textual methods, and LLM-based approaches on many evaluation metrics, validating the hypothesis that integrating relational and neighboring context via BERT enhances tail prediction accuracy. While FB15k-237 results suggest opportunities to optimize performance in dense networks, MuCo-KGC demonstrates exceptional adaptability and predictive precision, establishing itself as a robust solution for KGC.
\par
\noindent
\textbf{Efficiency and Parameter Optimization}
While our model's performance on FB15K-237 is less competitive in Hits@3 and Hits@10, it achieves this with a significantly reduced parameter count compared to SimKGC and DIFT. As shown in Table \ref{tab:prc}, our model uses 50\% fewer parameters than SimKGC and an impressive 98.43\% fewer parameters than DIFT. This highlights the model's efficiency and its potential for scalability in resource-constrained environments without heavily compromising performance.
\par
\noindent
\textbf{Limitations:} While our model demonstrates strong performance on datasets like WN18RR and CoDEx, its effectiveness on FB15k-237 is comparatively limited, likely due to the dataset’s higher relation cardinality, long-tail distribution, and increased complexity stemming from the removal of inverse triplets. These characteristics introduce noise, make pattern extraction more challenging, and amplify the difficulty of capturing robust embeddings for rare relations. Moreover, the model's reliance on contextual information may struggle with FB15k-237's heterogeneous and less semantically rich graph structure. Future work could focus on dynamic relation-aware sampling, enhanced multi-hop contexts tailored to high-diversity graphs, and loss functions addressing the long-tail distribution of entities and relations to improve robustness across more complex datasets.

\section{Ablation}
The ablation study (see Table \ref{tab:results}) assesses the contributions of the Head Context ($H_c$) and Relation Context ($R_c$) components in the MuCo-KGC model on FB15k-237 and WN18RR datasets.  The $H_c$-Only configuration, encapsulating the adjacent entities and relationships only, attains reasonable performance with an MRR of 0.310 and Hits@10 of 0.403 on FB15k-237, and an MRR of 0.420 and Hits@10 of 0.616 on WN18RR. The $R_c$-Only, which utilizes global relational patterns, shows inferior performance compared to $H_c$-Only, achieving an MRR of 0.280 and Hits@10 of 0.271 on FB15k-237, and an MRR of 0.321 and Hits@10 of 0.398 on WN18RR, highlighting the constraints of depending exclusively on global context. Evidently, each context independently contributes to enhance MuCo-KGC's performance.
%%%%%%%%%%%%%%%%%%%%%%%%%%%%%%%%%%%%%%%%%%%%%%%%%%%%%%%%%%%%%%%%%%%%%%
%Table \ref{tb:kgcca} given in Appendix \ref{sec:appenB}  presents our ablation studies on the FB15k-237 and WN18RR datasets, demonstrating the contribution of each component within the MuCo-KGC model. We progressively integrated components—head context, relation context, and relation context—leading to the complete MuCo-KGC model. The results reveal that each individual component enhances the model’s performance; however, none of the ablation models outperform the fully integrated MuCo-KGC. This confirms the strong effect of combining both head and relation contexts for optimal tail entity prediction.
%\hl{The uniqueness of the proposed MuCo-KGC approach is explicitly demonstrated in Table 2 through its unique integration of C3 (head context) and C5 (relationship context), which are missed by the current Knowledge Graph Completion (KGC) methods. While conventional models such as TransE, RotatE, and ConvE largely emphasize textual and relational data (C1 and C4), MuCo-KGC enhances contextual comprehension by integrating graph structure via neighbor information. Such dual focus enables MuCo-KGC to outperform alternative methods by utilizing more extensive and useful graph-based insights.}
%%%%%%%%%%%%%%%%%%%%%%%%%%%%%%%%%%%%%%%%%%%%%%%%%%%%%%%
\section{Conclusion}
The MuCo-KGC method offers an efficient approach for knowledge graph completion (KGC) by leveraging neighboring contextual information of the head entity and relation. It outperforms existing KGE-based and LLM-based methods across several benchmark datasets, with notable improvements in MRR and Hit@k metrics. Specifically, MuCo-KGC enhances Hit@1 performance on WN18RR and CoDEx-M by 6.88\% and 25.23\%, respectively.  Additionally, MuCo-KGC demonstrates an average improvement of 1.63\% on WN18RR, 3.77\% on CoDEx-S and 20.15\% on CoDEx-M in MRR metric, further showcasing its superior performance across diverse datasets. Ablation studies confirm the importance of incorporating contextual data for better performance. By eliminating the need for entity descriptions and negative triplet sampling, MuCo-KGC offers a scalable and accurate solution. Future work could explore the integration of long-range spatial information to enhance model performance in high-density datasets.

%%%%%%%%%%%%%%%%%%%%%%%%%%%%%%%%%%%%%%%%%%%%%%%%%%%%%%%%%%%%%%%%%%%%%%%%%%%%%%%%%%%%

%%%%%%%%%%%%%%%%%%%%%%%%%%%%%%%%%%%%%%%%%%%%%%%%

% ---- Bibliography ----
%
% BibTeX users should specify bibliography style 'splncs04'.
% References will then be sorted and formatted in the correct style.
%
\bibliographystyle{splncs04}
%\bibliography{mybibliography}

\end{document}